\documentclass[10pt,twocolumn,letterpaper]{article}

\usepackage{iccv}
\usepackage{times}
\usepackage{epsfig}
\usepackage{graphicx}
\usepackage{amsmath}
\usepackage{amssymb}
\usepackage{booktabs}
\usepackage{tabularx}
\usepackage[font=small,skip=5pt]{caption}
\usepackage[font=footnotesize,skip=5pt]{subcaption}
 \usepackage{url}

\usepackage{breakurl}
\usepackage{bbm}
\usepackage{multicol}
\usepackage{multirow}

\newcolumntype{Y}{>{\centering\arraybackslash\hsize=.2\hsize}X}

\DeclareMathOperator*{\argmin}{argmin}

\usepackage[breaklinks=true,bookmarks=false]{hyperref}

\iccvfinalcopy 

\def\iccvPaperID{3517} 

\ificcvfinal\fi

\begin{document}

\title{Calibrating Panoramic Depth Estimation for \\ Practical Localization and Mapping}

\author{Junho Kim\textsuperscript{1}, Eun Sun Lee\textsuperscript{1}, and Young Min Kim\textsuperscript{1, 2}
\and {\small \phantom{ }} \vspace{-1em}\\
\textsuperscript{1} {\small Dept. of Electrical and Computer Engineering, Seoul National University} \\
\textsuperscript{2} {\small Interdisciplinary Program in Artificial Intelligence and INMC, Seoul National University} \\
{\tt\small \{82magnolia, eunsunlee, youngmin.kim\}@snu.ac.kr}
}

\maketitle
\ificcvfinal\thispagestyle{empty}\fi

\begin{abstract}
The absolute depth values of surrounding environments provide crucial cues for various assistive technologies, such as localization, navigation, and 3D structure estimation.
We propose that accurate depth estimated from panoramic images can serve as a powerful and light-weight input for a wide range of downstream tasks requiring 3D information.
While panoramic images can easily capture the surrounding context from commodity devices, the estimated depth shares the limitations of conventional image-based depth estimation; the performance deteriorates under large domain shifts and the absolute values are still ambiguous to infer from 2D observations.
By taking advantage of the holistic view, we mitigate such effects in a self-supervised way and fine-tune the network with geometric consistency during the test phase.
Specifically, we construct a 3D point cloud from the current depth prediction and project the point cloud at various viewpoints or apply stretches on the current input image to generate synthetic panoramas.
Then we minimize the discrepancy of the 3D structure estimated from synthetic images without collecting additional data.
We empirically evaluate our method in robot navigation and map-free localization where our method shows large performance enhancements.
Our calibration method can therefore widen the applicability under various external conditions, serving as a key component for practical panorama-based machine vision systems.
Code is available through the following link: \url{https://github.com/82magnolia/panoramic-depth-calibration}.

\end{abstract}

\section{Introduction}

\begin{figure}[t]
\begin{center}
\includegraphics[width=\linewidth]{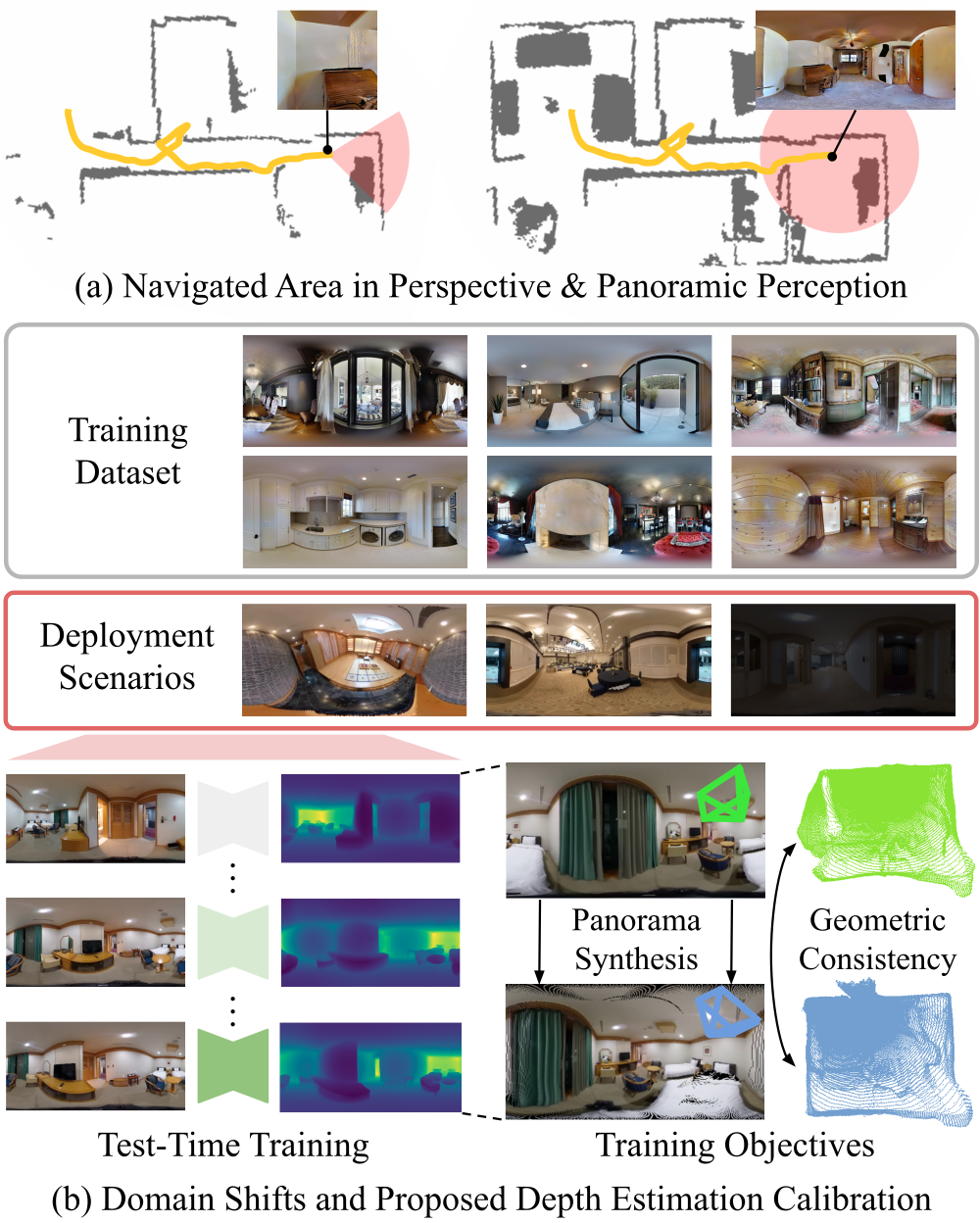}
\end{center}
\vspace{-1em}
   \caption{Motivation and overview of our approach. Panoramic perception enables efficient navigation due to the large field of view (top). Nevertheless, the performance drops due to the gaps between the training dataset with upright cameras in medium-sized rooms and the deployment scenarios with limited data and various domain shifts. The proposed solution suggests test-time training using geometric consistency to mitigate the gap (bottom). }
\label{fig:teaser}
\end{figure}

Acquiring depth maps of the surrounding environment is a crucial step for AR/VR and robotics applications, as the depth maps serve as building blocks for mapping and localization.
While dense LiDAR or RGB-D scanning~\cite{realsense_scan,matterport_scan,faro_scan,kinect_scan,velodyne_lidar} has been widely used for depth acquisition, the methods are often  computationally expensive or require costly hardware.
Panoramic depth estimation~\cite{acdnet,unifuse,bifuse,fredsnet,pano3d,omnifusion,omnidepth,omnidepth_synth}, on the other hand, enables quick and cost-effective depth computation.
It outputs a dense depth map from a single neural network inference given only $360^\circ$ camera input, which is becoming more widely accessible~\cite{Ricoh,insta360}.
Further, the large field of view of panoramic depth maps can model the comprehensive 3D context from a single image capture. 
The holistic view provides ample visual cues for robust localization, and allows efficient 3D mapping.
An illustrative example is shown in Figure~\ref{fig:teaser}a, where a robot navigation agent equipped with panorama view observes larger areas and builds more comprehensive grid map than the agent with perspective view when deployed for the same trajectory. 

While existing panoramic depth estimation methods can estimate highly accurate depth maps in trained environments~\cite{omnifusion,bifuse,unifuse,acdnet}, their performances often deteriorate when deployed in unseen environments with large domain gaps.
For example, as shown in Figure~\ref{fig:teaser}b, depth estimation networks trained on upright panorama images in medium-sized rooms perform poorly in images containing large camera rotation or captured in large rooms.
Such scenarios are highly common in AR/VR or robotics applications, yet it is infeasible to collect large amounts of densely annotated ground-truth data for panorama images or perform data augmentations to realistically and thoroughly cover all the possible adversaries.
Further, while numerous unsupervised domain adaptation methods have been proposed for depth estimation~\cite{chen2019crdoco,lopez2022desc,ssl_depth,feat_decomp_uda_depth}, most of them mainly consider sim-to-real gap minimization and require the labelled training dataset for adaptation which is infeasible for memory-limited applications.

In this paper, we propose a quick and effective calibration method for panoramic depth estimation in challenging environments with large domain shifts.
Given a pre-trained depth estimation network, our method applies test-time adaptation~\cite{tent,test_time_training,test_time_training_pp} on the network solely using objective functions derived from test data.
Conceptually, we are treating depth estimation networks as \textit{sensors} that output depth maps from images, which then makes the process similar to `calibration' in depth or LiDAR sensing literature for accurate measurements.
Our resulting scheme is flexibly applicable in either online or offline manner adaptation.
As shown in Figure~\ref{fig:teaser}c, the light-weight training \textit{calibrates} the network towards making more accurate predictions in the new environment.

\begin{figure}[t]
\begin{center}
\includegraphics[width=\linewidth]{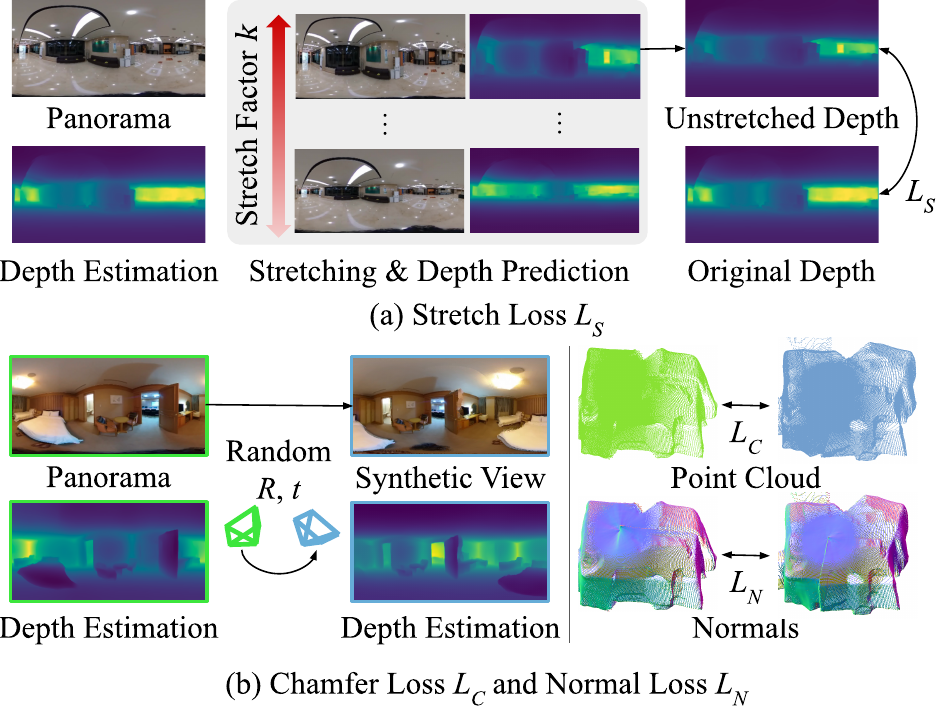}
\end{center}
\vspace{-1em}
   \caption{Description of the proposed test-time training objectives.}
\label{fig:loss}
\end{figure}

Our calibration scheme  consists of two key components that effectively utilize the holistic spatial context uniquely provided by panoramas.
First of all, our method operates using training objectives that impose geometric consistencies from novel view synthesis and panorama stretching.
To elaborate, as shown in Figure~\ref{fig:loss}, we leverage the full-surround 3D structure available from panoramic depth estimation and generate synthetic panoramas.
The training objectives then minimize the geometric discrepancy between depth estimations from the synthesized panoramas and the original view.
Second, we propose light-weight data augmentation to cope with offline scenarios where only a limited amount of test-time training data is available.
Specifically, we augment the test data by applying arbitrary pose shifts or synthetic stretches, similar to the techniques used for the training objectives.

Since our calibration method aims at adapting the network during the test phase using geometric consistencies, it is compute and memory efficient while being able to handle a wide variety of domain shifts.
Our method does not require the computational demands of additional network pre-training~\cite{test_time_training,pad}, or memory to store the original training dataset during adaptation~\cite{chen2019crdoco,ssl_depth,lopez2022desc,t2net}.
Nevertheless, our method shows large amounts of performance enhancements when tested in challenging domain shifts such as low lighting or room-scale change.
Further, due to the light-weight formulation, our method could easily be applied to numerous downstream tasks in localization and mapping.
We experimentally verify that our calibration scheme effectively improves performance in two exemplary tasks, namely map-free localization and robot navigation.
To summarize, our key contributions are as follows: (i) a novel test-time adaptation method for calibrating panoramic depth estimation, (ii) a data augmentation technique to handle low-resource adaptation scenarios, and (iii) an effective application of our calibration method on downstream mapping and localization tasks.
\section{Related Work}
\paragraph{Monocular Depth Estimation}
Following the pioneering work of Eigen et al.~\cite{Eigen2014DepthMP}, many existing works focus on developing neural network models that output depth maps from image input~\cite{ranftl2020towards,Ranftl2021,depth_1,depth_2,depth_3,depth_4}.
Recent approaches such as MiDAS~\cite{ranftl2020towards} or DPT~\cite{Ranftl2021} can make highly accurate depth predictions from images due to extensive training on large depth-annotated datasets~\cite{kitti,nyu_depth,megadepth}.
As a result, there have been numerous applications in localization and mapping that leverage monocular depth estimation.
For example, map-free visual localization~\cite{arnold2022map} localizes the camera position using maps built from monocular depth estimation, which is highly efficient compared to building a 3D map by Structure-from-Motion.
Another example is in robot navigation methods that directly estimate occupancy grid maps from input images~\cite{chaplot2020learning,occ_anticipation,Chaplot2020ObjectGN}, which could be implicitly regarded as  monocular depth estimation.

Compared to perspective images, monocular depth estimation using panorama images has been relatively understudied with the limited amount of available data.
While recent works~\cite{omnifusion,unifuse,fredsnet,acdnet,omnidepth,omnidepth_synth} have demonstrated accurate depth estimation in trained environments, their performance are known to deteriorate when tested in new datasets with varying lighting or depth distributions~\cite{pano3d}.
Such performance discrepancies are more noticeable for panoramic images since there are fewer depth-annotated images available compared to perspective images. 
One possible remedy is to re-project the panoramic image to multiple perspective images to create individual depth estimations and stitch the results together, as proposed by Rey-Area et al.~\cite{rey2022360monodepth} and Peng et al.~\cite{high_res}.
Although this may yield more robust depth estimation utilizing abundant previous frameworks for perspective images, the process involves fusing depth maps resulting from multiple neural network inferences, which is computationally expensive.
Our method takes a different direction of quickly calibrating the network to the new environment for robust panoramic depth estimation.
We demonstrate the effectiveness of our method on the aforementioned applications of depth estimation, namely map-free localization and robot navigation, to verify its practicality on downstream tasks.

\paragraph{Domain Adaptation for Depth Estimation}
A vast majority of domain adaptation methods target classification tasks~\cite{tent,test_time_training,test_time_training_pp,shot,sentry,pseudo_label_orig,mate}, and aim to minimize loss functions defined over the predicted class probabilities.
Existing methods could be categorized into those that only use the test data or those that require the original training dataset for adaptation.
For the former, pseudo-labelling methods~\cite{pseudo_label_orig}, masking-based methods~\cite{mask_ttt,mate}, batch normalization updating methods~\cite{batchnorm_update}, and fully test-time training methods~\cite{tent,tent_upgrade} impose self-supervised learning objectives to adapt to the test domain data.
On the other hand, methods belonging to the latter, namely unsupervised domain adaptation methods~\cite{sentry,uda_1,uda_2} and test time training methods with auxiliary task networks~\cite{test_time_training,test_time_training_pp,pad}, utilize the original training dataset to impose consistencies between the source and target domain predictions.

For depth estimation, most adaptation methods~\cite{chen2019crdoco,yen20223d,t2net,feat_decomp_uda_depth,ssl_depth} focus on the sim-to-real domain gap and apply techniques from pseudo-labelling or unsupervised domain adaptation.
Methods such as DESC~\cite{lopez2022desc} and 3D-PL~\cite{yen20223d} adapt using pseudo labels generated from additional semantic priors or style transferred predictions~\cite{adain,style_transfer_1,style_transfer_2,style_transfer_3}.
In contrast, unsupervised domain adaptation methods~\cite{t2net,chen2019crdoco,lopez2022desc} additionally utilize the depth-annotated training data and apply style transfer networks to learn a common feature representation between the source and target domain.
Compared to existing domain estimation methods for depth estimation, our calibration method can effectively handle a wider range of domain shifts and can perform light-weight online adaptation as no additional training data is required.
We extensively evaluate our method against existing domain adaptation techniques in Section~\ref{sec:exp}, where our method outperforms the tested baselines in various depth estimation scenarios.

\section{Method}
\label{sec:method}
Given a panoramic depth estimation network $F_\Theta(\cdot)$ trained on the source domain $\mathcal{S}$, the object of our calibration scheme is to adapt the network to a new, unseen target domain $\mathcal{T}$ during the test phase.
Our method could perform adaptation both in an online and offline manner: in the online case,  the network is simultaneously optimized and evaluated whereas in the offline case the network is first optimized using samples from the target domain and then evaluated with another set of target domain samples.
As shown in Figure~\ref{fig:loss}, our method leverages training objectives that impose geometric consistencies between the synthesized views generated from the full-surround depth predictions (Section~\ref{sec:training_obj}).
To further cope with offline adaptation scenarios where only a small number of images are available for training, we propose to apply data augmentation based on panorama synthesis (Section~\ref{sec:data_aug}).

\subsection{Test-Time Training Objectives}
\label{sec:training_obj}
Given a panorama image $I{\in} \mathbb{R}^{H{\times} W{\times} 3}$, the depth estimation network outputs a depth map $\hat{D}{=}F_\Theta(I) {\in} \mathbb{R}^{H{\times} W{\times} 1}$.
The test-time adaptation enforces consistencies between depth estimations of additional input images synthesized from the current predictions, and eventually achieves stable prediction under various environment setups.
The test-time training objective is given as follows,
\begin{equation}
\mathcal{L}=\mathcal{L}_{\text{S}}+\mathcal{L}_{\text{C}}+\mathcal{L}_{\text{N}},
\label{eq:total}
\end{equation}
where $\mathcal{L}_{\text{S}}$ is the stretch loss, $\mathcal{L}_{\text{C}}$ is the Chamfer loss, and $\mathcal{L}_{\text{N}}$ is the normal loss.

\paragraph{Stretch Loss}
Stretch loss aims to tackle the depth distribution shifts that commonly occur in panoramic depth estimation by imposing consistencies between depth predictions made at different panorama stretches.
Panoramic depth estimation models make large errors when confronted with images captured in scenes with drastic depth distribution changes~\cite{pano3d}.
The key intuition for stretch loss is to hallucinate the depth estimation network as if it is making predictions in a room with similar depth distributions as the trained source domain, through panoramic stretching shown in Figure~\ref{fig:loss}a.

The panorama stretching operation~\cite{stretch_1,stretch_2} warps the input panorama $I$ to a panorama captured from the same 3D scene but stretched along the $x, y$ axes by a factor of $k$.
For a panorama image $I$ this could be expressed as follows,
\begin{equation}
    \mathcal{S}_\text{img}^k(I)[u, v]=I[u, \frac{H}{\pi}\arctan(\frac{1}{k}\tan(\frac{\pi v}{H}))],
\end{equation}
where $I[u, v]$ is the color value at coordinate $(u, v)$ and $\mathcal{S}_\text{img}^k(\cdot)$ is the $k$-times stretching function for images.
A similar operation could be defined for depth maps, namely
\begin{equation}
    \mathcal{S}_\text{dpt}^k(D)[u, v]=\kappa(v) D[i, \frac{H}{\pi}\arctan(\frac{1}{k}\tan(\frac{\pi v}{H}))],
\end{equation}
where $\kappa(v){=}\sqrt{k^2\sin^2(\pi v/H){+}\cos^2(\pi v/H)}$ is the correction term to account for the depth value changes due to stretching.

Stretch loss enforces depth predictions made at large scenes to follow predictions made at \textit{contracted} scenes (using $k < 1$) and for those at small scenes to follow predictions made at \textit{enlarged} scenes (using $k > 1$).
The distinction between large and small scenes is made by thresholding on the average depth value using thresholds $\delta_1, \delta_2$.
Formally, this could be expressed as follows,
\begin{equation}
     \mathcal{L}_\text{S} {=}
\left\{\arraycolsep=1.8pt
	\begin{array}{ll}
	   {\sum_{k \in \mathcal{K}_s}}\|\hat{D} {-} \mathcal{S}^{1/k}_\text{dpt}(F_\Theta(\mathcal{S}^k_\text{img}(I)))\|_2  & \mbox{if $\text{avg}(\hat{D}) < \delta_1$} \\
	   {\sum_{k \in \mathcal{K}_l}}\|\hat{D} {-} \mathcal{S}^{1/k}_\text{dpt}(F_\Theta(\mathcal{S}^k_\text{img}(I)))\|_2  & \mbox{if $\text{avg}(\hat{D}) > \delta_2$} \\
  0 & \mbox{otherwise,}
	\end{array}
\right.
\label{eq:stretch}
\end{equation}
where $\text{avg}(\hat{D})$ is the pixel-wise average for the depth map $\hat{D}=F_\Theta(I)$, and $\mathcal{K}_l=\{\sigma, \sigma^2\}, \mathcal{K}_s=\{1/\sigma, 1/\sigma^2\}$ are the stretch factors used for contracting and enlarging panoramas.
In our implementation we set $\delta_1{=}1, \delta_2{=}2.5, \sigma{=}0.8$.

\paragraph{Chamfer and Normal Loss}
While stretch loss guides depth predictions to have coherent scale, chamfer and normal loss encourages depth predictions to be geometrically consistent at a finer level.
The loss functions operate by generating synthetic views at small random pose perturbations from the original viewpoint, and minimizing discrepancies between depth predictions made at synthetic views and the original view.

First, the Chamfer loss minimizes the Chamfer distance between depth predictions made at different poses.
Given a panoramic depth map $D$, let $\mathcal{B}(D): \mathbb{R}^{H\times W \times 1}\rightarrow \mathbb{R}^{HW \times 3}$ denote the back-projection function that maps each pixel's depth values $D[u, v]$ to a point in 3D space $D[u, v]*S[u,v]$ where $S[u,v]\in\mathbb{R}^3$ is a point on the unit sphere corresponding to the panorama image coordinate $(u,v)$.
Further, let $\mathcal{W}(I, D; R, t)$ denote the warping function that outputs an image rendered at an arbitrary pose $R, t$, as shown in Figure~\ref{fig:loss}b.
Then, the Chamfer loss is given as follows,
\begin{equation}
    \mathcal{L}_\text{C}=\sum_{\mathbf{x}\in \mathcal{B}(\hat{D})}\min_{\mathbf{y}\in \mathcal{B}(D_\text{warp})}\|\Tilde{R}\mathbf{x}+\Tilde{t} - \mathbf{y}\|_2^2,
\label{eq:chamfer}
\end{equation}
where $D_\text{warp}=F_\Theta(\mathcal{W}(I, \hat{D}; \Tilde{R}, \Tilde{t}))$ is the depth prediction made from the warped image at a randomly chosen pose $\Tilde{R}, \Tilde{t}$ near the origin.
We choose $\Tilde{R}$ to be a random rotation around the z-axis and $\Tilde{t}$ as a random translation sampled from $[-0.5, 0.5]^3$.

Second, the normal loss imbues an additional layer of geometric consistency by aligning normal vectors of the depth maps.
Let $\mathcal{N}(\mathbf{x}): \mathbb{R}^{3}\rightarrow \mathbb{R}^{3}$ denote the normal estimation function that uses ball queries around the point $\mathbf{x}$ to compute the normal vector~\cite{normal_1,normal_2}.
The normal loss is then given as follows,

\begin{equation}
    \mathcal{L}_\text{N}=\sum_{\mathbf{x} \in \mathcal{B}(\hat{D})} (\Tilde{R}\mathcal{N}(\mathbf{x}) \cdot (\Tilde{R}\mathbf{x}+\Tilde{t} - \argmin_{\mathbf{y}\in \mathcal{B}(D_\text{warp})}\|\Tilde{R}\mathbf{x}+\Tilde{t}-\mathbf{y}\|_2))^2
\end{equation}

Conceptually, the normal loss minimizes the distance between planes spanning from points in the original depth map $\hat{D}$ against the nearest points in the warped image's depth map $D_\text{warp}$.
Note this is similar in spirit to loss functions used in point-to-plane ICP~\cite{point_to_plane}.
We further verify the effectiveness of each loss function in Section~\ref{sec:exp}.

\subsection{Data Augmentation}
We propose a data augmentation scheme that increases the number of test-time training data for offline adaptation scenarios where only a small number of target domain data is available in a real-world deployment.
For example, a robot agent may need to quickly adapt to the new environment after observing a few samples, or AR/VR applications may want to quickly build an accurate 3D map of the environment using a small set of images.

Key to our augmentation scheme is the panorama synthesis from stretching and novel view generation.
Given a single panorama image $I$ and its associated depth prediction $\hat{D}=F_\Theta(I)$, the augmentation scheme $\mathcal{A}(I, \hat{D})$ for generating a new panorama is given as follows,
\begin{equation}
     \mathcal{A}(I, \hat{D}) {=}
\left\{\arraycolsep=1.8pt
	\begin{array}{ll}
            \mathcal{W}(I, \hat{D}; \Tilde{R}, \Tilde{t}) & \mbox{if $\text{avg}(\hat{D}) \in [\delta_1, \delta_2]$ } \\
            \mathcal{S}^k_\text{img}(I) & \mbox{otherwise,} \\
	\end{array}
\right.
\label{eq:augment}
\end{equation}
where $\Tilde{R}, \Tilde{t}$ are random poses sampled near the origin and $k$ is randomly sampled from $\mathcal{U}(\sigma^2, \sigma)$ if $\text{avg}(\hat{D}) > \delta_2$ and $\mathcal{U}(1/\sigma, 1/\sigma^2)$ if $\text{avg}(\hat{D}) < \delta_1$.
The values for $\delta_1,\delta_2,\sigma$ are identical to those used in Section~\ref{sec:training_obj}.
Conceptually, our augmentation scheme generates novel views at random poses if the average depth values are within a range $[\delta_1, \delta_2]$ and applies stretching otherwise, where the scene size determines the stretch factor.
Despite the simple formulation, our augmentation scheme enables test-time adaptation only using a small number of image data (at the extreme case, even with a \textit{single} training sample), where we further demonstrate its effectiveness by illustrating its applications in Section~\ref{sec:app}.

\label{sec:data_aug}
\section{Applications}
\label{sec:app}
In this section, we show applications of our panoramic depth calibration on two downstream tasks, robot navigation, and map-free localization.

\begin{figure}[t]
\begin{center}
   \includegraphics[width=0.98\linewidth]{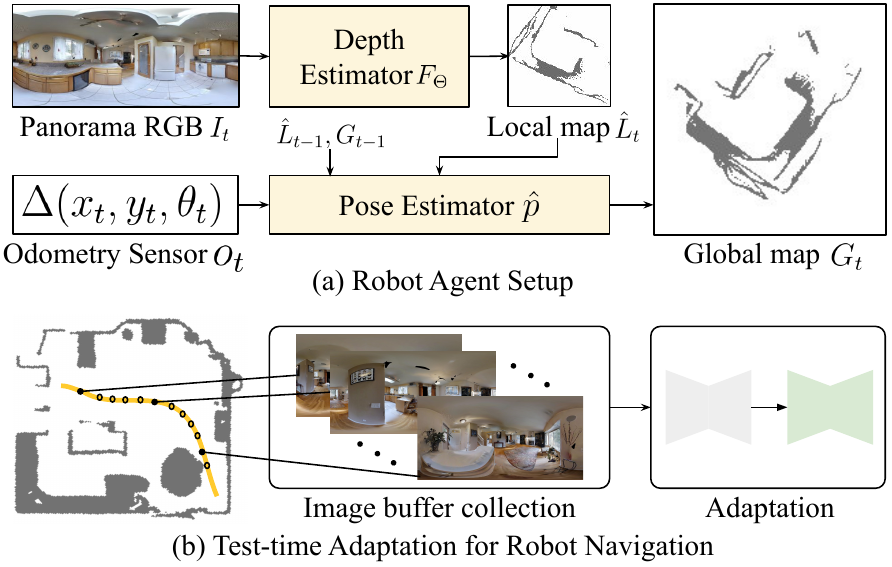}
\end{center}
   \caption{Robot agent with panoramic perception (top) and application of panoramic depth calibration on robot navigation task (bottom).}
\label{fig:nav}
\end{figure}

\subsection{Robot Navigation}
\label{sec:robot}
\paragraph{Navigation Agent Setup} 
We assume a navigation agent equipped with a panorama camera and noisy odometry sensor, similar to the setup of recently proposed navigation agents~\cite{chaplot2020learning,occ_anticipation,Chaplot2020ObjectGN}.
As shown in Figure~\ref{fig:nav}a, for each time step $t$ the navigation agent first creates a local 2D occupancy grid map $\hat{L}_t$ based on the depth estimation results from the panorama, namely $F_\Theta(I_t)$.
Then, the pose estimation network observes the previous and current local map $(\hat{L}_{t-1}, \hat{L}_t)$ along with the noisy odometry sensor reading $o_t$ to produce a pose estimate $\hat{p}_t=C_\Phi(\hat{L}_{t-1}, \hat{L}_t, o_t)$.
The pose estimate is further used to stitch the local map $L_t$ against the previous global map $G_{t-1}$ to form an updated global map $G_t$.
Finally, the policy network takes the global grid map and the current image observation as input to output an action policy, namely $a_t=P_\Psi(G_t, I_t)$, where the possible actions are to move forward by 0.25m or turn left or right by $10^\circ$.

\paragraph{Depth Calibration for Robot Navigation}
We begin each navigation episode by applying our test-time training to calibrate the panoramic depth estimates from a small number of visual observations collected.
As shown in Figure~\ref{fig:nav}b, the agent caches the first $N_\text{fwd}$ panoramic views seen after it makes a forward action.
Then, using the data augmentation from Section~\ref{sec:data_aug} $N_\text{aug}$ times for each cached image, the agent performs test-time training with the $N_\text{fwd}\times N_\text{aug}$ set of images.
Once the calibration is completed, the agent uses the updated depth estimation network to create the global map and compute policy for the subsequent steps remaining in the episode.
Note that the calibration process for navigation terminates very quickly, with the total number of training steps for each episode being smaller than $300$ steps.
Nevertheless, the quick calibration results in significant performance improvements for various downstream navigation tasks, which is further verified in Section~\ref{sec:exp}.

\subsection{Map-Free Visual Localization}
\label{sec:loc}
\paragraph{Localization Process Overview} 
First introduced by Arnold et al.~\cite{arnold2022map}, map-free visual localization aims at finding the camera pose with respect to a 3D scene where the conventional Structure-from-Motion (SfM) mapping process is omitted, hence the name `map-free'.
Instead, the 3D scene is represented using a 3D point cloud obtained from monocular depth estimation, which in turn greatly reduces the computational burden required for obtaining SfM maps.

\begin{figure}[t]
\begin{center}
   \includegraphics[width=\linewidth]{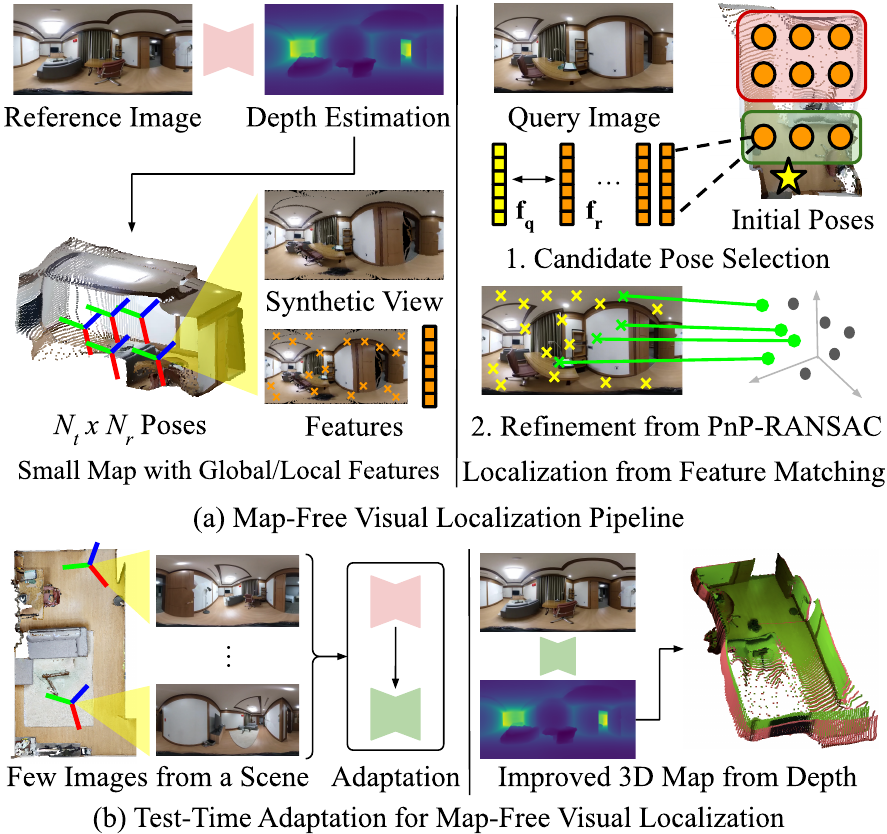}
\end{center}
\vspace{-1em}
   \caption{Description of map-free localization task (top) and its test-time adaptation pipeline (bottom). }
\label{fig:loc}
\end{figure}

We adapt the original map-free localization framework designed for perspective cameras to panoramas, and validate our calibration scheme on the task.
As shown in Figure~\ref{fig:loc}, given a single reference image $I_\text{ref}$ and its associated depth prediction $\hat{D}_\text{ref}=F_\Theta(I_\text{ref})$, map-free localization initiates with generating a 3D map from the depth map, namely $\mathcal{B}(\hat{D}_\text{ref})$.
Then we generate synthetic panoramas for the $N_t \times N_r$ poses $\{(R_i, t_i)\}$ and extract global/local feature descriptors, where $N_t$ translations and $N_r$ rotations uniformly sampled from the bounding box of $\mathcal{B}(\hat{D}_\text{ref})$.

During localization, global and local descriptors are first similarly extracted for the query image $I_q$.
Then, the top-$K$ poses from the pool of $N_t \times N_r$ poses are chosen whose euclidean distances of the global descriptors are closest to that of the query image $\mathbf{f}_q$.
The selected poses are further ranked with local feature matching using SuperGlue~\cite{sarlin2020superglue}, where the candidate pose with the largest number of matches is refined for the final prediction.
Here, for each local feature match between the query image and synthetic view we retrieve the corresponding 3D point from the point cloud $\mathcal{B}(\hat{D}_\text{ref})$ and apply PnP-RANSAC, as shown in Figure~\ref{fig:loc}.

\paragraph{Depth Calibration for Map-Free Localization}
For each 3D scene, we assume only a small handful of images (between $1\sim 5$) are available for adaptation, to reflect AV/VR application scenarios where the user wants to quickly localize in a new environment.
Depth calibration is then applied to fine-tune the depth estimator, where we increase the number of training samples using data augmentation similar to robot navigation.
After calibration, the modified network is applied to create a 3D map from an arbitrary reference image captured from the same environment, which could then be used for localizing new query images.

\section{Experimental Results}
\label{sec:exp}
We first evaluate how our calibration scheme enhances depth prediction (Section~\ref{sec:exp_depth_estim}).
We then validate its effect on the aforementioned applications, namely robot navigation (Section~\ref{sec:exp_robot_nav}) and map-free visual localization (Section~\ref{sec:exp_map_free}).

\begin{figure}[t]
\begin{center}
   \includegraphics[width=0.93\linewidth]{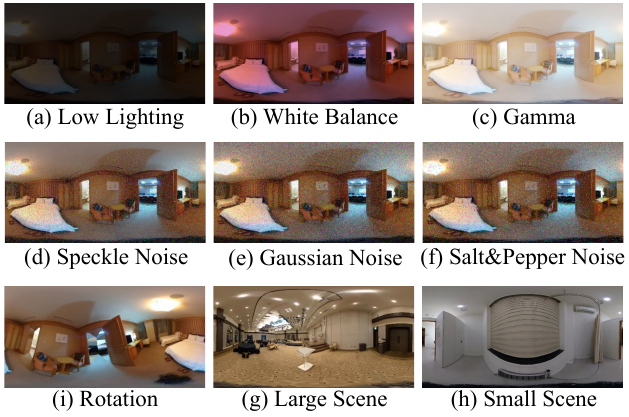}
\end{center}
\vspace{-1em}
   \caption{Visualization of domain changes.}
\label{fig:domain}
\end{figure}

\vspace{-1em}
\paragraph{Implementation Details}
We implement our method using PyTorch~\cite{paszke2019pytorch}, and use the pre-trained UNet from Albanis et al.~\cite{pano3d} as the original network for adaptation.
The network is trained using the depth-annotated panorama images from the Matterport3D dataset~\cite{matterport}.
For test-time training, we optimize the loss function from Equation~\ref{eq:total} using Adam~\cite{kingma2014adam} for 1 epoch, with a learning rate of $10^{-4}$ and batch size of 4.
In all our experiments, we use the RTX 2080Ti GPU for acceleration.
Additional details about the implementation is deferred to the supplementary material.

\vspace{-1em}
\paragraph{Datasets}
Unlike the common practice of panoramic depth estimation~\cite{omnifusion,unifuse,bifuse} where the train/test splits are created from the same dataset, we consider entirely different datasets from the training dataset for evaluation.
Specifically, we use the Stanford 2D-3D-S dataset~\cite{armeni2017joint} and OmniScenes~\cite{kim2021piccolo} dataset for the depth estimation and map-free localization experiments, and the Gibson~\cite{xia2018gibson} dataset equipped with the Habitat simulator~\cite{habitat19iccv} for robot navigation experiments.
Both Stanford 2D-3D-S and OmniScenes datasets contain a diverse set of 3D scenes, with 1413 panoramas captured from 272 rooms for Stanford 2D-3D-S dataset and 7614 panoramas captured from 18 rooms for OmniScenes.
The Gibson dataset contains 14 scenes for the validation split, which is used on top of the Habitat simulator~\cite{habitat19iccv} to evaluate various robot navigation tasks.

\vspace{-1em}
\paragraph{Baselines}
As our task has not been studied in previous works, we adapt existing test-time adaptation and unsupervised domain adaptation methods to panoramic depth estimation and implement six baselines.

The four test-time adaptation baselines only use the test data for adaptation.
Tent~\cite{tent} only updates the batch normalization layer during adaptation, where we implement a variant that minimizes the loss function from Equation~\ref{eq:total}.
Flip consistency-based approach (FL) inspired by Li et al~\cite{li2021self} enforces the depth predictions between the original and flipped image to be similar.
Mask consistency-based approach (MA) inspired by Mate~\cite{mate} enforces depth consistency against a randomly masked panorama image.
Pseudo Labeling (PS)~\cite{pseudo_label_depth} imposes losses against the pseudo ground-truth depth map by averaging predictions made from multiple rotated panoramas.

The two unsupervised domain adaptation methods additionally use the labeled source domain dataset for adaptation, where we use AdaIN~\cite{adain} to perform style transfer between the source and target domain images.
Vanilla T\textsuperscript{2}Net minimizes the discrepancy between the depth predictions of the source domain image transferred to the target domain and the ground truth.
CrDoCo~\cite{chen2019crdoco} additionally makes the target domain predictions to follow the predictions of the target-to-source transferred images.
We provide detailed expositions of each baseline on the supplementary material.

\begin{figure}[t]
\begin{center}
\includegraphics[width=1\linewidth]{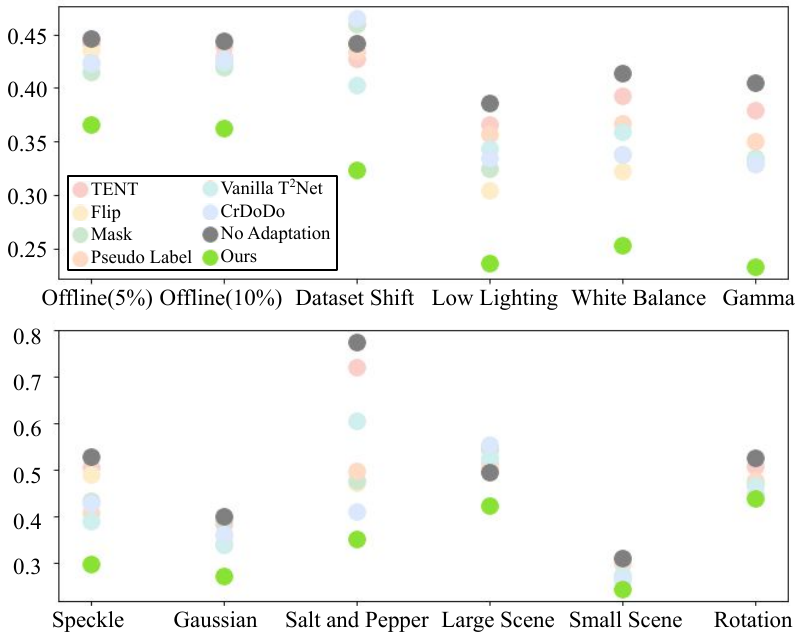}
\end{center}
\vspace{-1em}
   \caption{Plot of online adaptation result. The result of our method compared to the baselines with various domain changes (top) and image noises (bottom). }
\label{fig:bar_plot}
\end{figure}

\begin{figure}[t]
\begin{center}
\includegraphics[width=1.0\linewidth]{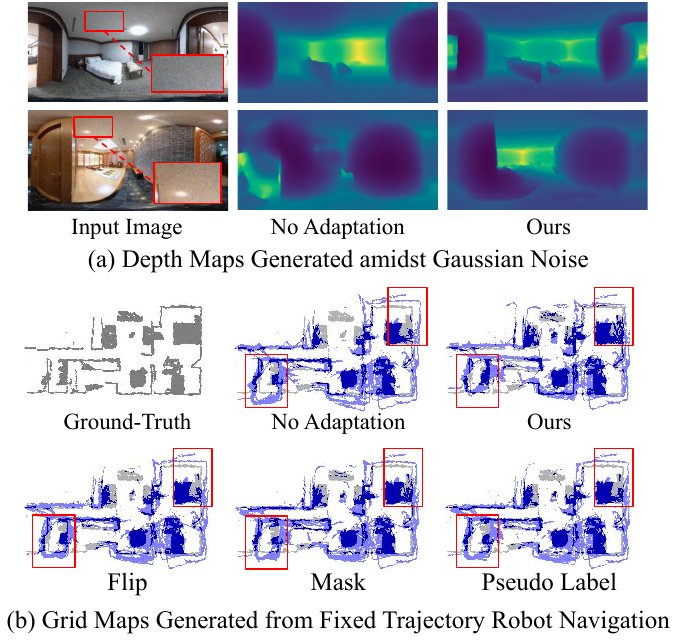}
\end{center}
\vspace{-1em}
   \caption{Qualitative result of depth maps 
   (top) and grid maps from navigation task (bottom). }
\label{fig:qual}
\end{figure}

\subsection{Depth Estimation}
\label{sec:exp_depth_estim}
\paragraph{Online Adaptation}
As shown in Figure~\ref{fig:domain} we evaluate our method on 10 target domains: generic dataset change, 3 global lighting changes (image gamma, white balance, average intensity), 3 image noises (gaussian, speckle, salt \& pepper), 3 geometric changes (scene scale change to large/small scenes, camera rotation).
For scene scale change we use the large rooms manually selected from the evaluation datasets, and for other domains we use the scikit-image library~\cite{scikit_image} to generate the image-level changes.  
We provide additional implementation details about the domain setups in the supplementary material.

Figure~\ref{fig:bar_plot} summarizes the mean absolute error (MAE) of various adaptation methods aggregated from the Stanford2D-3D-S~\cite{armeni2017joint} and OmniScenes~\cite{kim2021piccolo} datasets.
We report the full evaluation results in the supplementary material.
Our method outperforms the baselines across all tested domain shifts, with more than 10cm decrease in MAE in most shifts.
The loss functions presented in Section~\ref{sec:training_obj} thus enables effective depth network calibration.
For large scene adaptation, the tested baselines fail to make sufficient performance improvements, whereas our method can largely reduce the error via stretch loss.
In addition, note that our method can perform adaptation even in photometric domain shifts such as speckle noise or white balance change, despite the geometry-centric formulation.
The multi-view consistency imposed by normal and Chamfer loss ensures the network to make more robust depth predictions amidst these adversaries.
A few exemplary depth visualizations are shown in Figure~\ref{fig:qual}, where our online calibration results in depth maps with more accurate depth scales and detail preservation. 
We report the full results against other depth estimation metrics in the supplementary material.

\vspace{-1em}
\paragraph{Offline Adaptation}
We additionally experiment with offline adaptation scenarios, where the depth network is first trained on a small set of images and then tested on a held-out set.
To cope with the data scarcity during training, we apply data augmentation for all the tested methods with $N_\text{aug}=10$.
For evaluation, we apply our calibration method separately for each room in the Stanford2D-3D-S~\cite{armeni2017joint} and OmniScenes~\cite{kim2021piccolo} datasets, where the panoramas captured for each room are split for training and testing.
Figure~\ref{fig:bar_plot} shows the adaptation results, where the results are reported after using $5\%$ or $10\%$ of panoramas at each room for training.
In both evaluations, our method incurs large amounts of performance enhancement while outperforming all the tested baselines.

\begin{table}[]
\centering
\resizebox{0.8\columnwidth}{!}{%
\begin{tabularx}{1.08\columnwidth}{l|cccc}
\toprule
Method & MAE & Abs. Rel. & RMSE & Sq. Rel.\\ \midrule
No Adatation & 0.4343 & 0.1949 & 0.6025 & 0.1428 \\
Ours w/o Stretch Loss & 0.3650 & 0.1500 & 0.5450 & 0.1001 \\
Ours w/o Chamfer Loss & 0.3260 & 0.1468 & 0.4819 & 0.1115 \\
Ours w/o Normal Loss & 0.3373 & 0.1512 & 0.5022 & 0.0953 \\
Ours w/o Augmentation & 0.3972 & 0.1790 & 0.5566 & 0.1247 \\
Ours & \textbf{0.3192} & \textbf{0.1432} & \textbf{0.4683} & \textbf{0.0906} \\
\bottomrule
\end{tabularx}
}
\caption{Ablation study of key components of our calibration scheme. `Abs. Rel.' and `Sq. Rel.' denote the absolute and squared relative error from Eigen et al.~\cite{Eigen2014DepthMP}.}
\label{tab:abl}
\end{table}

\paragraph{Ablation Study} 
To further validate the effectiveness of the various components in our calibration scheme, we perform an ablation study on the offline adaptation setup.
We use the OmniScenes~\cite{kim2021piccolo} dataset for evaluation and use $10\%$ of panoramas in each room for training and the rest for testing.
As shown in Table~\ref{tab:abl}, omitting any one of the loss functions leads to suboptimal performance.
In addition, the data augmentation scheme incurs a large amount of performance boost, which indicates that despite its simplicity, data augmentation plays a crucial role in data-scarce offline adaptation scenarios.

\begin{table}[t]
\begin{subtable}{\linewidth}
    \centering
    \resizebox{0.9\linewidth}{!}{
    \begin{tabular}{l|cc|cc}
        \toprule
     & \multicolumn{2}{c}{Exploration} & \multicolumn{2}{|c}{Point Goal} \\
     \midrule
    Method & \multicolumn{1}{c}{Exp. Ratio} & \multicolumn{1}{c|}{Coll. Rate} & \multicolumn{1}{c}{Success Rate} & \multicolumn{1}{c}{Coll. Rate} \\
            \midrule
    No Adaptation & 0.8835 & 0.2793 & 0.4107 & 0.3864 \\
    Flip Consistency & 0.9027 & \textbf{0.2305} & 0.5000 & 0.3516 \\
    Mask Consistency & 0.8758 & 0.2677 & \textbf{0.4643} & 0.3460 \\
    Pseudo Labeling & 0.8701 & 0.8701 & 0.3393 & 0.4058 \\
    Ours & \textbf{0.9288} & 0.2352 & \textbf{0.4643} & \textbf{0.3135} \\
    \bottomrule
    \end{tabular}
    }
\caption{Exploration and Point Goal Navigation}
\end{subtable}

\smallskip

\begin{subtable}{\linewidth}
    \centering
    \resizebox{0.9\linewidth}{!}{
    \begin{tabular}{l|cc|cc}
        \toprule
         & \multicolumn{2}{c}{Localization} & \multicolumn{2}{|c}{Mapping} \\
         \midrule
        Method & \multicolumn{1}{c}{$t$-Error (m)} & \multicolumn{1}{c}{$R$-Error ($^\circ$)} & \multicolumn{1}{|c}{Cmf. Dist.} & \multicolumn{1}{c}{MAE} \\
        \midrule
        No Adaptation & 0.1450 & 10.2731 & 0.1959 & 0.4963 \\
        Flip Consistency & 0.1329 & 10.1872 & 0.1687 & 0.4743 \\
        Mask Consistency & 0.1336 & \textbf{10.1356} & 0.1585 & 0.4637 \\
        Pseudo Labeling & 0.1347 & 10.3247 & 0.1735 & 0.4711 \\
        Ours & \textbf{0.1177} & 10.2044 & \textbf{0.1558} & \textbf{0.4453} \\
        \bottomrule
        \end{tabular}    
    }
\caption{Localization and Mapping under Fixed Trajectory}

\end{subtable}
\caption{Robot navigation evaluation against existing methods.}

\label{tab:nav_eval}
\end{table}
\begin{table}[]
\centering
\setlength{\tabcolsep}{6pt}
\resizebox{\columnwidth}{!}{%
\begin{tabularx}{1.35\columnwidth}{l|YYYY}
\toprule
Method & \begin{tabular}[c]{@{}c@{}}$t$-error \\ (m)\end{tabular} & \begin{tabular}[c]{@{}c@{}}$R$-error \\ ($^\circ$)\end{tabular} & \begin{tabular}[c]{@{}c@{}}Accuracy \\ (0.1m, $5^\circ$)\end{tabular} & \begin{tabular}[c]{@{}c@{}}Accuracy \\ (0.2m, $10^\circ$)\end{tabular} \\ \midrule
No Adatation & 0.16 & 0.91 & 0.32 & 0.62 \\
Flip Consistency& 0.14 & 0.79 & 0.36 & 0.67 \\
Mask Consistency& 0.12 & 0.88 & 0.41 & 0.77 \\
CrDoCo & 0.11 & 1.00 & 0.46 & 0.78 \\
Ours & \textbf{0.09} & \textbf{0.87} & \textbf{0.52} & \textbf{0.86}\\
\bottomrule
\end{tabularx}
}
\caption{Map-Free visual localization compared against the baselines. Note that the translation and rotation error thresholds for calculating accuracy is denoted as $(d\text{ m}, \theta^\circ)$.}
\label{tab:loc_eval}
\end{table}

\subsection{Robot Navigation}
\label{sec:exp_robot_nav}
We consider three tasks for evaluating robot navigation using panoramic depth estimation, following prior works~\cite{chaplot2020learning, self_sup_slam, moda}: point goal navigation, exploration, and simultaneous localization and mapping (SLAM) from a fixed robot trajectory. 
First, point goal navigation aims to navigate the robot agent towards a goal specified from the agent's starting location, e.g. ``move to the location 5m forward and 10m right from the origin". 
Second, the objective of exploration is to explore the given 3D scene as much as possible under a fixed number of action steps.
Finally, the SLAM task evaluates the accuracy of the occupancy grid map and pose estimates under a fixed robot trajectory.
We use 4 random starting points at each of 14 scenes in the Gibson~\cite{xia2018gibson} dataset totaling 56 episodes per task, and set the maximum number of action steps to $500$.

Table~\ref{tab:nav_eval} compares the robot navigation tasks against three baselines (Flip consistency, Mask consistency, and Pseudo Labeling) where our method outperforms the baselines in most metrics.
For exploration, our calibration scheme results in largest exploration areas and rates while attaining a small collision rate, which is the total collision divided by the total number of action steps.
A similar trend is present for point goal navigation, where our agent attains the highest success rate with the smallest number of collisions.
Note that the success rate is computed as the ratio of navigation episodes where the robot reached within 0.2m of the designated point goal.
Finally for fixed-trajectory SLAM, our method exhibits higher localization and mapping accuracy than its competitors.
The translation error for localiztion drops largely after adaptation, while the rotation error is similar across all the baselines which is due to the $360\deg$ field-of-view that makes rotation estimation fairly accurate even prior to localization.
On the mapping size, our method attains the smallest 2D Chamfer distance and image error metrics (MAE) measured between the estimated global map and the ground-truth.
In addition, as shown in Figure~\ref{fig:qual} the grid maps resulting from our method best aligns with the ground truth when compared against the maps from the baselines.
Thus, the training objectives along with the light-weight augmentation enables quick and effective adaptation for various navigation tasks.

\subsection{Map-Free Visual Localization}
Similar to the offline evaluation explained in Section~\ref{sec:exp_depth_estim}, for each room in the OmniScenes~\cite{kim2021piccolo} dataset we select $5\%$ of the panorama images for test-time training and the rest for evaluating localization.
Then, we treat each evaluation image as the reference image $I_\text{ref}$ from Section~\ref{sec:loc} and generating a 3D map via depth estimation.
To finally evaluate localization we query 10 images that are captured within 2m of each reference image, where we use the dataset's 6DoF pose annotations to determine the criterion. 

Table~\ref{tab:loc_eval} shows the localization performance compared against three baselines (Flip consistency, Mask consistency, and CrDoCo~\cite{chen2019crdoco}).
Following prior works in visual localization~\cite{piccolo,hierarchical_scene,posenet}, we report the median translation and rotation errors along with accuracy where a prediction is considered correct if its translation and rotation error is below a designated threshold.
Our method outperforms the baselines in both tested datasets, with almost a $20\%$ increase in accuracy.
The geometry correction of our method as shown in Figure~\ref{fig:qual} leads to more accurate PnP-RANSAC solutions, which in turn results in enhanced localization performance.

\label{sec:exp_map_free}

\section{Conclusion}
We propose a simple yet effective calibration scheme for panoramic depth estimation.
Domain shifts between training and deployment is a critical problem in panoramic depth estimation as a slight change in the camera pose or lighting can incur large performance drops, while such adversaries are common in practical application scenarios.
We introduce three training objectives along with an augmentation scheme to mitigate the domain shifts, where the key idea is to impose geometric consistency via panorama synthesis from random pose perturbations and stretching.
Further, our experiments show that the light-weight formulation can largely improve performance on downstream applications in mapping and localization.
Backed by the plethora of recent advancements in panoramic depth estimation, we project our calibration scheme to function as a key ingredient for practical full-surround depth sensing.

\paragraph{Acknowledgements}
This work was supported by the National Research Foundation of Korea(NRF) grant funded by the Korea government(MSIT) (No. RS-2023-00208197) and Samsung Electronics Co., Ltd. Young Min Kim is the corresponding author.

\appendix
\renewcommand\thetable{\thesection.\arabic{table}}    
\setcounter{table}{0}
\renewcommand\thefigure{\thesection.\arabic{figure}}    
\setcounter{figure}{0}
\newcommand{\Hquad}{\hspace{0.2em}} 

\pagenumbering{gobble}

\section{Implementation Details}
\subsection{Loss Functions for Test-Time Training}
As explained in Section~\textcolor{red}{3.1}, our calibration method involves fine-tuning the depth estimation network using three training objectives.
In this section we explain how each objectives are implemented, along with the detailed hyperparameter setups.

\paragraph{Stretch Loss}
The goal of stretch loss is to mitigate the domain gap that occurs from depth scale changes in small or large scenes as shown in Figure~\ref{fig:supp_depth}.
The loss minimizes the difference between the stretched depth values against the original depth prediction, namely
\begin{equation}
     \mathcal{L}_\text{S} {=}
\left\{\arraycolsep=1.8pt
	\begin{array}{ll}
	   {\sum_{k \in \mathcal{K}_s}}\|\hat{D} {-} \mathcal{S}^{1/k}_\text{dpt}(F_\Theta(\mathcal{S}^k_\text{img}(I)))\|_2  & \mbox{if $\text{avg}(\hat{D}) < \delta_1$} \\
	   {\sum_{k \in \mathcal{K}_l}}\|\hat{D} {-} \mathcal{S}^{1/k}_\text{dpt}(F_\Theta(\mathcal{S}^k_\text{img}(I)))\|_2  & \mbox{if $\text{avg}(\hat{D}) > \delta_2$} \\
  0 & \mbox{otherwise,}
	\end{array}
\right.
\label{eq:stretch}
\end{equation}
where $\text{avg}(\hat{D})$ is the pixel-wise average for the depth map $\hat{D}=F_\Theta(I)$, and $\mathcal{K}_l=\{\sigma, \sigma^2\}, \mathcal{K}_s=\{1/\sigma, 1/\sigma^2\}$ are the stretch factors used for contracting and enlarging panoramas.
We use the publicly available codebase from Sun et al.~\cite{stretch_1} to implement the stretching operations $\mathcal{S}^{k}_\text{img}, \mathcal{S}^{k}_\text{dpt}$, and set $\delta_1{=}1, \delta_2{=}2.5, \sigma{=}0.8$.

\paragraph{Chamfer and Normal Loss}
Along with stretch loss that enforces scale consistency, Chamfer loss and Normal loss impose fine-grained geometric consistency.
Both losses operate by creating synthetic views rendered at random translations and rotations, where we adapt the codebase of Zioulis et al.~\cite{omnidepth_synth} to implement the rendering operation.
First, given a panorama image $I$, Chamfer loss is given as follows,
\begin{equation}
    \mathcal{L}_\text{C}=\sum_{\mathbf{x}\in \mathcal{B}(\hat{D})}\min_{\mathbf{y}\in \mathcal{B}(D_\text{warp})}\|\Tilde{R}\mathbf{x}+\Tilde{t} - \mathbf{y}\|_2^2,
\label{eq:chamfer}
\end{equation}
where $\mathcal{B}(\hat{D})$ is the point cloud created from the original depth prediction and $\mathcal{B}(D_\text{warp})=\mathcal{B}(F_\Theta(\mathcal{W}(I, \hat{D}; \Tilde{R}, \Tilde{t})))$ is the point cloud from the depth prediction made at a synthesized view from a randomly chosen pose $\Tilde{R}, \Tilde{t}$ near the origin.
We implement the Chamfer loss using the {\small\texttt{chamfer\_distance}} function from the PyTorch3D library~\cite{pytorch3d}.

The Normal loss imbues an additional level of geometric consistency by aligning the normal vectors between the predictions for the original and synthetic views.
The Normal loss is  defined as follows,
\begin{equation}
    \mathcal{L}_\text{N}=\sum_{\mathbf{x} \in \mathcal{B}(\hat{D})} (\Tilde{R}\mathcal{N}(\mathbf{x}) \cdot (\Tilde{R}\mathbf{x}+\Tilde{t} - \argmin_{\mathbf{y}\in \mathcal{B}(D_\text{warp})}\|\Tilde{R}\mathbf{x}+\Tilde{t}-\mathbf{y}\|_2))^2
\end{equation}
where $D_\text{warp}$ is a depth map from an arbitrary translation and rotation $\Tilde{R}, \Tilde{t}$, and $\mathcal{N}(\mathbf{x})$ is the normal vector at point $\mathbf{x}$.
We implement the Normal loss using the {\small\texttt{estimate\_pointcloud\_normals}} function from PyTorch3D~\cite{pytorch3d} and set the number of ball queries as $15$.

\subsection{Robot Navigation}
We implement the navigation agent similar to Active Neural SLAM~\cite{chaplot2020learning} and use the Habitat simulator~\cite{habitat19iccv} for evaluating the application of our calibration method on robot navigation.
As explained in Section \textcolor{red}{4.1}, we consider an agent that receives a panorama image and noisy odometry sensor reading as inputs and draws an occupancy grid map.
While the original implementation of Active Neural SLAM~\cite{chaplot2020learning} trains the entire set of navigation modules end-to-end, we use the pre-trained depth estimation network $F_\Theta$ from Albanis et al.~\cite{pano3d} and only train the policy network $P_\Psi$ and pose estimator network $C_\Phi$ on the Gibson training split~\cite{habitat19iccv}.
For depth calibration, the augmentation factor is $N_\text{aug}=10$ in all our experiments. 
The test-time training process caches images as shown in Figure \textcolor{red}{3}, using the first $N_\text{fwd}=25$ images for the exploration and SLAM tasks and $N_\text{fwd}=3$ images for the point goal navigation task. 
We use a larger number of cached images for exploration and SLAM tasks as the tasks generally take longer steps compared to the point goal task. 

\subsection{Map-Free Localization}
We implement a structure-based localization method~\cite{sarlin2019coarse} based on the setup explained in Section \textcolor{red}{4.2}, The query image $I_q$ is localized against a small 3D map $\mathcal{B}(\hat{D}_\text{ref})$ created from a reference image $I_\text{ref}$.
To elaborate, we first generate synthetic panoramas at $N_t \times N_r$ poses and global/local features.
During localization, the features are compared against the query image features to choose candidate poses and refine them using PnP-RANSAC~\cite{epnp,ransac}.
In our implementation, we use NetVLAD~\cite{netvlad} for global features and SuperPoint~\cite{superpoint} for local features, which are both widely used for visual localization~\cite{arnold2022map,sarlin2019coarse}.
Further, we set the number of translations as $N_t=100$ and rotations as $N_r=8$.
For rotations, we assume that the gravity direction is known and generate $N_r$ rotation matrices by only varying the yaw angle values.

\section{Additional Experimental Details and Results}
\subsection{Baseline Comparison in Depth Estimation}
In Section \textcolor{red}{5.1}, we establish comparisons against various baselines for depth estimation amidst domain changes.
For evaluation we use the Stanford 2D-3D-S~\cite{stanford2d3d} and OmniScenes~\cite{piccolo} datasets.
Note for OmniScenes we use the `turtlebot' split as other splits contain moving human hands and bodies whose ground-truth depth values are not available.
Below we elaborate on the domain and baseline setups, and provide the additional experimental results of depth estimation.
\subsubsection{Domain Setup}
\label{sec:dom_setup}
\paragraph{Online Adaptation}
For online adaptation, we evaluate our method in 10 domain shifts using the Stanford 2D-3D-S~\cite{stanford2d3d} and OmniScenes~\cite{piccolo} datasets. The domain shifts shown in Figure \textcolor{red}{5} are implemented as follows:
\begin{itemize}
    \item \textbf{Dataset Shift:} We do not apply any additional transformations to the images. The images from the tested datasets are directly used for evaluation.
    \item \textbf{Low Lighting:} We lower each pixel intensity by 25\%.
    \item \textbf{White Balance:} We apply the following transformation matrix to the raw RGB color values: $\begin{pmatrix}
    0.7 & 0 & 0\\
    0 & 0.9 & 0\\
    0 & 0 & 0.8
    \end{pmatrix}$.
    \item \textbf{Gamma:} We set the image gamma to 1.5.
    \item \textbf{Speckle:} We use the {\small\texttt{random\_noise}} function from the scikit-image library, where we set the speckle noise variance parameter to $0.06$.
    \item \textbf{Gaussian:} We use the same library as in speckle noise, where we set the Gaussian noise variance parameter to $0.005$.
    \item \textbf{Salt and Pepper:} We use the same library as in speckle noise, where we randomly perturb $0.5\%$ of the image pixels.
    \item \textbf{Large Scene:} For OmniScenes~\cite{piccolo}, we select the {\small\texttt{wedding}}, {\small\texttt{lounge}}, and {\small\texttt{lobby}} scenes and for Stanford 2D-3D-S~\cite{stanford2d3d} we select all rooms labelled as {\small\texttt{hallway}} and {\small\texttt{auditorium}}.
    \item \textbf{Small Scene:} For OmniScenes~\cite{piccolo}, we select the {\small\texttt{bride\_room}}, {\small\texttt{makeup\_room}}, and {\small\texttt{pyebaek}} scenes and for Stanford 2D-3D-S~\cite{stanford2d3d} we select all rooms labelled as {\small\texttt{pantry}}, {\small\texttt{WC}}, {\small\texttt{storage\_room}} and {\small\texttt{copy\_room}}.
    \item \textbf{Rotations:} We apply a random rotation on the test     images with yaw angles sampled from $\mathcal{U}(-\pi, \pi)$, roll angles sampled from $\mathcal{U}(-\pi / 8, \pi / 8)$, and pitch angles sampled from $\mathcal{U}(-\pi / 8, \pi / 8)$.
\end{itemize}
\paragraph{Offline Adaptation}
For offline adaptation, we separately evaluate depth estimation in each room for OmniScenes~\cite{piccolo} and Stanford 2D-3D-S~\cite{stanford2d3d}.
Specifically, we select 5\% or 10\% of panorama images in each room for training, and evaluate using the remaining images.
Note that for the Stanford 2D-3D-S dataset, many rooms contain less than 20 panoramas, which means that often only a \textit{single} image is used for adaptation.
To cope with data scarcity, we apply data augmentation from Section \textcolor{red}{3.2} for $N_\text{aug}=20$ times in the 5\% case and $N_\text{aug}=10$ times in the 10\% case to increase the test-time training data.

\subsubsection{Baseline Setup}

For evaluating our calibration method, we test against seven baselines in the main paper.
Here we elaborate on the implementations of each baseline, along with three additional baselines which we make detailed comparisons in Section~\ref{sec:full_exp}.
In the offline setup, all baselines are trained for a single epoch to ensure fair comparison.

\paragraph{Tent and Batch Normalization Statistics Update}
Introduced by Wang et al.~\cite{tent}, Tent proposes to only train the affine parameters from the batch normalization layer during test-time training.
Similar to Tent, Schneider et al.~\cite{batchnorm_update} propose to only update the batch normalization statistics during adaptation.
We adapt both baselines to our setup while for Tent we modify the original entropy-based training objective to our training objective in Equation \textcolor{red}{1} to accommodate for the task change from classification to depth prediction.

\paragraph{Flip Consistency}
Originally developed for self-supervised visual odometry~\cite{li2021self}, flip consistency imposes consistency against the flipped input image.
Formally, the baseline minimizes the flip consistency loss given as follows,
\begin{equation}
    \mathcal{L}_\text{flip}=\|F_\Theta(\mathcal{T}_\text{flip}(I)) - \mathcal{T}_\text{flip}(F_\Theta(I))\|_2,
\end{equation}
where $\mathcal{T}_\text{flip}(\cdot)$ is the horizontal flipping operation.

\paragraph{Mask Consistency}
Inspired from Mate~\cite{mate}, the mask consistency baseline operates by first creating a randomly masked image and imposing depth consistency against the original prediction.
Let $\Tilde{M} \in \mathbb{R}^{H \times W}$ denote the random mask generated for each test sample.
Then, the baseline is trained with the following objective,
\begin{equation}
    \mathcal{L}_\text{mask} = \|\Tilde{M} \circ F_\Theta(I) - \Tilde{M} \circ F_\Theta(\Tilde{M} \circ I) \|_2,
\end{equation}
where $\circ$ is the member-wise product operation.
We implement the random masking operation by first splitting the input panorama into $N_h \times N_w$ square patches and randomly discarding 10\% of the patches, where we set $N_h=4, N_w=8$.

\paragraph{Photometric Consistency}
Similar to the loss functions often used for self-supervised depth estimation~\cite{godard2017unsupervised,unsup_depth_1}, the photometric consistency baseline imposes consistency between the synthesized view using depth estimation results and the original view.
The baseline first generates a synthetic panorama located at translation $\Tilde{t}$ and rotation $\Tilde{R}$ from the origin, namely $I_\text{warp}=\mathcal{W}(I, \hat{D}; \Tilde{R}, \Tilde{t})$.
Then, the baseline minimizes the following loss,
\begin{equation}
    \mathcal{L}_\text{photo}=\|I-\mathcal{W}(I_\text{warp}, D_\text{warp}; \Tilde{R}^{-1}, \Tilde{t}^{-1})\|_2,
\end{equation}
where $D_\text{warp}$ is the depth estimation using $I_\text{warp}$.

\paragraph{Pseudo Labelling}
First introduced by Lee et al.~\cite{pseudo_label_orig}, the pseudo labelling baseline creates a pseudo ground-truth by aggregating depth predictions made at various rotated panoramas.
Formally, the baseline minimizes the following objective,
\begin{equation}
    \mathcal{L}=\|F_\Theta(I) - \frac{1}{K}\sum_{k=1}^{K}\mathcal{T}_\text{rot}(F_\Theta(\mathcal{T}_\text{rot}(I, \frac{2\pi}{K})), -\frac{2\pi}{K}), \|_2
\end{equation}
where $\mathcal{T}_\text{rot}(I, 2\pi/K)$ denotes the horizontal rotation of the input panorama $I$ by $2\pi / K$ rad.
In all our experiments we set $K=4$.

\paragraph{Unsupervised Domain Adaptation}
We consider three unsupervised domain adaptation baselines: vanilla T\textsuperscript{2}Net, CrDoCo~\cite{chen2019crdoco}, and feature consistency~\cite{ssl_depth}.
All three baselines use the original source domain dataset during adaptation, which is the Matterport3D~\cite{matterport} dataset in our implementation.
For each test sample, we randomly sample an image and ground-truth depth pair $(\Tilde{I}_\text{src}, \Tilde{D}_\text{src})$ and use them for adapting the network to the new domain.
In addition, the baselines utilize a style transfer network $F_\text{style}(I, I_\text{ref})$ that transforms the input panorama $I$ to match the style of the reference panorama $I_\text{ref}$.
We implement $F_\text{style}$ based on AdaIN~\cite{adain}, which is a widely used method for style transfer.

First, vanilla T\textsuperscript{2}Net imposes consistency between the style transferred depth prediction and the original depth prediction, namely
\begin{equation}
    \mathcal{L}_\text{T\textsuperscript{2}Net}=\|F_\Theta(F_\text{style}(I_\text{src}, I)) - D_\text{src}\|_2.
\end{equation}
Note that here the source domain image $I_\text{src}$ is transformed to match the style of the target domain image $I$.
While the original T\textsuperscript{2}Net imposes an additional set of adversarial losses based on GANs~\cite{t2net}, we omit those losses and hence the baseline is named \textit{vanilla} T\textsuperscript{2}Net.
As our setup does not target a single transition from sim-to-real but a wide range of domain shifts and the number of test data is highly limited, it is infeasible to train a set of generators and discriminators for each domain shift.

CrDoCo~\cite{chen2019crdoco} builds upon vanilla T\textsuperscript{2}Net and imposes an additional cross-domain consistency loss, namely
\begin{equation}
    \mathcal{L}_\text{CrDoCo}=\mathcal{L}_\text{T\textsuperscript{2}Net} + \|F_\Theta(F_\text{style}(I, I_\text{src})) - F_\Theta(I)\|_2.
\end{equation}
Conceptually, the new loss of CrDoCo transforms the target domain image to match the source domain image. It enforces the original depth prediction $F_\Theta(I)$ to follow the transformed prediction.

Finally, the feature consistency baseline~\cite{ssl_depth} imposes an additional loss to impose consistency between the intermediate activations of the depth prediction network. 
This could be expressed as follows,
\begin{align}
\mathcal{L}_\text{feat}&=\mathcal{L}_\text{CrDoCo} + \|F_\Theta^\text{inter}(F_\text{style}(I, I_\text{src})) - F_\Theta^\text{inter}(I)\|_2 \nonumber \\
&+ \|F_\Theta^\text{inter}(F_\text{style}(I_\text{src}, I)) - F_\Theta^\text{inter}(I_\text{src})\|_2,
\end{align}
where $F_\Theta^\text{inter}$ is the intermediate layer activations of the depth estimation network $F_\Theta$. 

\paragraph{Ground-Truth Training}
To measure the upper-bound performance, we finally consider a baseline that uses the ground-truth data from the target domain.
Specifically, the ground-truth training baseline minimizes the following loss,
\begin{equation}
    \mathcal{L}_\text{gt}=\|F_\Theta(I)-D_\text{gt}\|_2,
\end{equation}
where $D_\text{gt}$ is the ground-truth depth map for image $I$.

\begin{table}[t]
\centering
\begin{subtable}{0.5\linewidth}
    \centering
    \resizebox{1.07\linewidth}{!}{
    \begin{tabular}{l|cc}
        \toprule
        Method &  Area (m\textsuperscript{2}) & Collisions \\
        \midrule
        No Adaptation & 28.4169 & 7822\\
        Flip Consistency & 28.8958 & \textbf{6440}\\
        Mask Consistency & 28.8796 & 7481\\
        Pseudo Labelling & 27.9863 & 7324\\
        Ours & \textbf{30.5147} & 6589\\
        \bottomrule
    \end{tabular}
    }
\caption{Exploration}
\end{subtable}
\Hquad
\begin{subtable}{0.45\linewidth}
    \centering
    \resizebox{0.8\linewidth}{!}{
    \begin{tabular}{l|cccc}
        \toprule
        Method &  PSNR \\
        \midrule
        No Adaptation & 7.2667 \\
        Flip Consistency & 7.4840\\
        Mask Consistency & 7.5850\\
        Pseudo Labelling & 7.5000\\
        Ours & \textbf{7.8404} \\
        \bottomrule
    \end{tabular}
    }
\caption{SLAM w/ Fixed Trajectory}
\end{subtable}
\vspace{-0.5em}
\caption{Additional metrics for the exploration and SLAM task in robot navigation.}
\label{table:supp_navigation}

\end{table}

\subsubsection{Full Experimental Results}
\label{sec:full_exp}
We report the full experimental results for depth estimation, where we present results from i) Stanford 2D-3D-S~\cite{stanford2d3d}, ii) OmniScenes~\cite{piccolo}, and iii) the aggregated total results.
Here we compare our calibration method against the baselines using six metrics, namely mean absolute error (MAE), absolute relative difference (Abs. Rel.), squared relative difference (Sq. Rel.), root mean squared error (RMSE), log root mean squared error (RMSE (Log)), and inlier ratio.
Each metric is defined as follows:
\begin{itemize}
    \item \textbf{MAE:} $\sum_{u,v}\frac{|D[u,v]-D_\text{gt}[u,v]|}{H*W}$
    \item \textbf{Abs. Rel.:} $\sum_{u,v}\frac{|D[u,v]-D_\text{gt}[u,v]|}{H*W*D_\text{gt}[u,v]}$
    \item \textbf{Sq. Rel.:} $\sum_{u,v}\frac{|D[u,v]-D_\text{gt}[u,v]|^2}{H*W*D_\text{gt}[u,v]}$ 
    \item \textbf{RMSE:} $\sqrt{\sum_{u,v}\frac{|D[u,v]-D_\text{gt}[u,v]|^2}{H*W}}$
    \item \textbf{RMSE (Log):} $\sqrt{\sum_{u,v}\frac{|\log {D[u,v]}-\log{D_\text{gt}[u,v]}|^2}{H*W}}$
    \item \textbf{Inlier Ratio:} $\sum_{u,v}\frac{1}{H*W}\mathbbm{1}\{\max{(\frac{D[u,v]}{D_\text{gt}[u,v]},\frac{D_\text{gt}[u,v]}{D[u,v]})} {<} \lambda\}$, where 
    $\lambda$ is a pre-defined inlier threshold.
\end{itemize}

\begin{figure}[t]
\begin{center}
\includegraphics[width=1\linewidth]{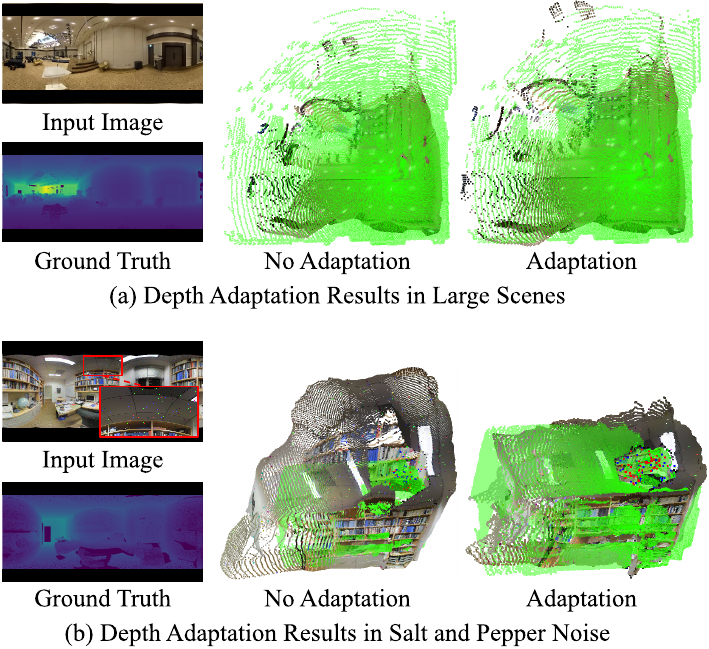}
\end{center}
\vspace{-1em}
   \caption{Qualitative visualization of depth estimation before and after adaptation. We overlay the ground-truth depth values in green.}
\label{fig:supp_depth}
\end{figure}

As shown in Table \textcolor{red}{B.4} to Table \textcolor{red}{B.39}, our method outperforms most of the baselines in all tested metrics.
We also display visualizations of the depth values before and after adaptation in Figure~\ref{fig:supp_depth}, where our adaptation scheme can largely alleviate the quality deterioration from domain shifts. 
The depth estimation results suggest that our calibration method can serve as an effective enhancement scheme in practical depth estimation scenarios.
\begin{table}[]
\centering
\resizebox{0.6\columnwidth}{!}{%
\begin{tabularx}{0.75\columnwidth}{l|cc}
\toprule
Loss Function & Room 5 & Wedding Hall \\ \midrule
No Adaptation & 0.4839 & 1.1704 \\
Flip & 0.4605 & 1.2746 \\
Mask & 0.3644 & 1.3302 \\
Photometric & 0.4889 & 1.1177 \\
Pseudo Labelling & 0.4435 & 1.2935 \\
Normal & \textbf{0.2699} & 1.3437 \\
Stretch & 0.4741 & \textbf{0.9167} \\
Chamfer & 0.2917 & 1.3537 \\
\bottomrule
\end{tabularx}
}
\caption{Mean absolute error of various test-time training loss functions measured from rooms in OmniScenes~\cite{piccolo}.}
\label{tab:loss_comp}
\end{table}

\subsubsection{Loss Function Comparison}
We additionally establish comparisons between the loss functions used in our calibration method (normal, stretch, Chamfer) against the loss functions used in the baselines.
To this end we run a small experiment where we evaluate offline adaptation on two rooms in OmniScenes~\cite{piccolo} (Room 5 and Wedding Hall) using 10\% of the available images for training and the rest for testing.
Here Room 5 exemplifies the `dataset shift' case, whereas Wedding Hall exemplifies the `large scene' case explained from Section~\ref{sec:dom_setup}.
We additionally apply an augmentation for each loss function by $N_\text{aug}=10$ and measure the mean absolute error on the test samples.
As shown in Table~\ref{tab:loss_comp}, both normal loss and Chamfer loss outperform the other loss functions in Room 5 while stretch loss shows large improvements in the wedding hall scene.
The fine-grained geometric consistencies from normal and Chamfer loss, along with the scale consistencies imposed from stretch loss enable our calibration method to function in a wide variety of depth estimation scenarios.

\subsection{Robot Navigation}

\paragraph{Experimental Setup} 
In all our experiments, we set the maximum number of steps per episode to $500$, while the point goal navigation task terminates whenever the agent stops within 0.2m of its estimated goal position.
Also, note that the Chamfer distance metric shown in Table \textcolor{red}{2b} is measured by treating the occupied regions in the grid map as a 2D point cloud.
For the SLAM task under fixed trajectory shown in Table \textcolor{red}{2b}, we collect the trajectories by having the `No Adaptation' robot agent to explore in each episode for 500 steps.

\begin{figure}[t]
\begin{center}
\includegraphics[width=1\linewidth]{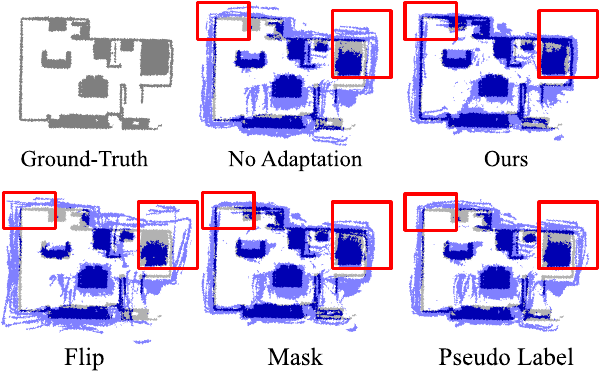}
\end{center}
\vspace{-1em}
   \caption{Qualitative result of grid maps from navigation task. We display the ground-truth map (grey) and the estimated grid map (blue) from the same sequence of actions.}
\label{fig:supp_map}
\end{figure}

\paragraph{Additional Results}
We report additional metrics and visualizations for the exploration and SLAM tasks in Table~\ref{table:supp_navigation} and Figure~\ref{fig:supp_map}.
First, we show the average explored area and the total number of collisions occurred during exploration. Our method achieves the highest explored area while exhibiting a lower collision count than most of the competing methods.
In addition, we display the PSNR between the estimated grid maps and the ground truth, where our method attains the highest grid map similarity.
This is further verified through the qualitative samples in Figure~\ref{fig:supp_map}, where the grid map generated using our calibration scheme best aligns with the ground-truth. 
Therefore, our method could effectively function in various robot navigation tasks to enhance their performance in challenging deployment scenarios.

\begin{table}[]
\centering
\setlength{\tabcolsep}{13pt}
\begin{subtable}{\linewidth}
\centering
\resizebox{0.95\columnwidth}{!}{%
\begin{tabularx}{1.38\columnwidth}{l|YYYY}
\toprule
\multirow{2}{*}{Method} & $t$-error & $R$-error & Accuracy & Accuracy \\ 
&(m) & ($^\circ$)& (0.1m,$5^\circ$)& (0.2m,$10^\circ$)\\
\midrule
No Adatation & 0.09 & \textbf{0.54} & 0.54 & 0.96 \\
Flip Consistency & 0.09 & 0.58 & 0.57 & 0.98 \\
Mask Consistency & 0.07 & 0.58 & 0.66 & 0.98 \\
CrDoCo & 0.07 & 0.61 & 0.70 & \textbf{0.99} \\
Ours & \textbf{0.05} & 0.60 & \textbf{0.83} & 0.96 \\
\bottomrule
\end{tabularx}
}
\caption{Query images within 1m of reference image}
\end{subtable}

\smallskip

\begin{subtable}{\linewidth}
\centering
\resizebox{0.95\columnwidth}{!}{%
\begin{tabularx}{1.38\columnwidth}{l|YYYY}
\toprule
\multirow{2}{*}{Method} & $t$-error & $R$-error & Accuracy & Accuracy \\ 
&(m) & ($^\circ$)& (0.1m,$5^\circ$)& (0.2m,$10^\circ$)\\
\midrule
No Adatation & 0.26 & 1.35 & 0.20 & 0.42 \\
Flip Consistency & 0.22 & 1.26 & 0.24 & 0.46 \\
Mask Consistency & 0.19 & 1.43 & 0.29 & 0.51 \\
CrDoCo & 0.19 & 1.41 & 0.31 & 0.53 \\
Ours & \textbf{0.15} & \textbf{1.13} & \textbf{0.38} & \textbf{0.60} \\
\bottomrule
\end{tabularx}
}
\caption{Query images within 3m of reference image}
\end{subtable}

\caption{Additional results of map-free visual localization compared against the baselines in OmniScenes~\cite{piccolo}. Note that the translation and rotation error thresholds for calculating accuracy is denoted as $(d\text{ m}, \theta^\circ)$.}.
\label{tab:supp_loc_1}
\end{table}

\subsection{Map-Free Localization}
As explained in Section \textcolor{red}{5.3}, we evaluate map-free localization by querying 10 images that are captured within a distance threshold $\delta=\text{2m}$ of each reference image.
Also, for test-time training we augment the data by $N_\text{aug}=20$ and cope with data scarcity.
In this section we additionally report results for $\delta=\text{1, 3m}$.
Table~\ref{tab:supp_loc_1} shows the localization results at various distance thresholds, where our method outperforms the baselines in most tested setups.
Along with robot navigation, our method demonstrates large amounts of performance enhancements in map-free localization, where the refined geometry of the depth maps play a crucial role for accurate localization.

{\small
\bibliographystyle{ieeetr}
\bibliography{main}
}

\end{document}